# A Novel Neuromorphic Processors Realization of Spiking Deep Reinforcement Learning for Portfolio Management

Seyyed Amirhossein Saeidi[1], Forouzan Fallah[2], Soroush Barmaki[3], Hamed Farbeh[4]
*Department of Computer Engineering, Amirkabir University of Technology, Tehran, Iran*
[1]sahsaeedi@aut.ac.ir, [2]foruzan.fallah@aut.ac.ir, [3]sr.barmaki@aut.ac.ir, [4]farbeh@aut.ac.ir

*Abstract*— The process of continuously reallocating funds into financial assets, aiming to increase the expected return of investment and minimizing the risk, is known as *portfolio management*. Processing speed and energy consumption of portfolio management have become crucial as the complexity of their real-world applications increasingly involves high-dimensional observation and action spaces and environment uncertainty, which their limited onboard resources cannot offset. Emerging neuromorphic chips inspired by the human brain increase processing speed by up to 1000 times and reduce power consumption by several orders of magnitude. This paper proposes a spiking deep reinforcement learning (SDRL) algorithm that can predict financial markets based on unpredictable environments and achieve the defined portfolio management goal of profitability and risk reduction. This algorithm is optimized for *Intel's Loihi neuromorphic* processor and provides 186x and 516x energy consumption reduction is observed compared to the competitors, respectively. In addition, a 1.3x and 2.0x speed-up over the high-end processors and GPUs, respectively. The evaluations are performed on cryptocurrency market between 2016 and 2021 the benchmark.

## I. INTRODUCTION

The continuous decision-making process of allocating a certain amount of capital to several different financial investment products for maximizing the return on risk control is called *portfolio management* (PM) [1], [2]. Traditional PM methods are divided into 4 sections: "*Follow-Winner*", "*Follow-the-Loser*", "*Pattern matching*" and "*Meta-Learning*" [3]. The performance of these methods was not sufficient and their results depended on the validity of the models in different markets. Because of the complexity of PM process, it causes four major problems: a) Repeat cycles, b) Ignoring real-world factors in the first stages, c) Simplified trading rules by human transactions, and d) Optimizing the inefficient trading system [4].

Deep machine learning methods were developed for financial market transactions in recent years [5], [6]. Although these methods were performing better than traditional methods, their accuracy in predicting the future was not enough due to their dependence on the financial market history. Recently, several methods have been presented based on deep reinforcement learning with relatively acceptable results [7]–[12]. One of the best methods is offered by [12]. Using a deep learning network called *Policy*, this method predicts an estimate of the capital distribution between different stocks instead of predicting the future of the financial market. This method has two main drawbacks.

a) Due to the depth of the Policy network, the reinforcement learning process (RL) is limited and has no improvement with increasing learning time and b) Due to the use of deep channels in the Policy network, feature extraction is not done well. As a result, it limits the learning process [7].

With the advancement of technology and the increasing number of IoT devices, the interest in solving real-world problems on devices such as smartphones and embedded systems has increased [13]. High processing power consumption of deep reinforcement learning (DRL) methods has made it impossible to use these methods on devices with limited power budget, e.g., smartphones [14]. Another problem with the PM is the high inference time due to the complex models in powerful processors such as the GPU [4]. Reducing the inference time requires shrinking the policy grid, which may be due to the ample space of the issue not being processed and the algorithm not reaching the optimal policy [15].

Energy efficiency is one of the main advantages of Spiking Neural Networks (SNNs). SNNs are an emerging architecture inspired by the human brain. In this architecture, neurons are calculated asynchronously and communicate through discrete events called spikes [16].

We show how SNNs create a low-power solution on neuromorphic processors such as *Intel's Loihi*. Most DRL methods perform the learning process based on maximizing reward. Despite the biologically acceptable nature of this learning rule, SNNs suffer from catastrophic forgetfulness and lack of policy evaluation, which limits its ability to learn policies in complex environments [17]. The DRL method uses reply memory to evaluate policies to overcome forgetfulness. This led to the question of the extent to which we can use the possibilities of DRL methods in SNN-based methods. Recent efforts have made it possible to use reply memory alongside the SNN-based method [18].

Recently, a policy-gradient algorithm has been proposed to train an SNN for policy learning [19]. However, this algorithm is for a discrete action space with constraints on the dimensions of the ongoing problem. Another factor that limits the learning of policy by SNNs is high dimensionality.

Interestingly, the abstracting of the brain topology and its computational principles have recently led to the design of new SNNs that exhibit human-like behaviors and improve performance [20], [21]. A key feature in the brain associated with efficient computation is the use of neurons to display information from sensory stimuli to output signals [22]. Early studies on population coding have shown its ability to display better stimuli, which has led to recent success in training SNNs to reduce high-dimensional complexity [23], [24]. The effectiveness shown in population coding opens up prospects for developing population-based SNNs that can lead to optimal policies on ongoing high-dimensional problems [15].

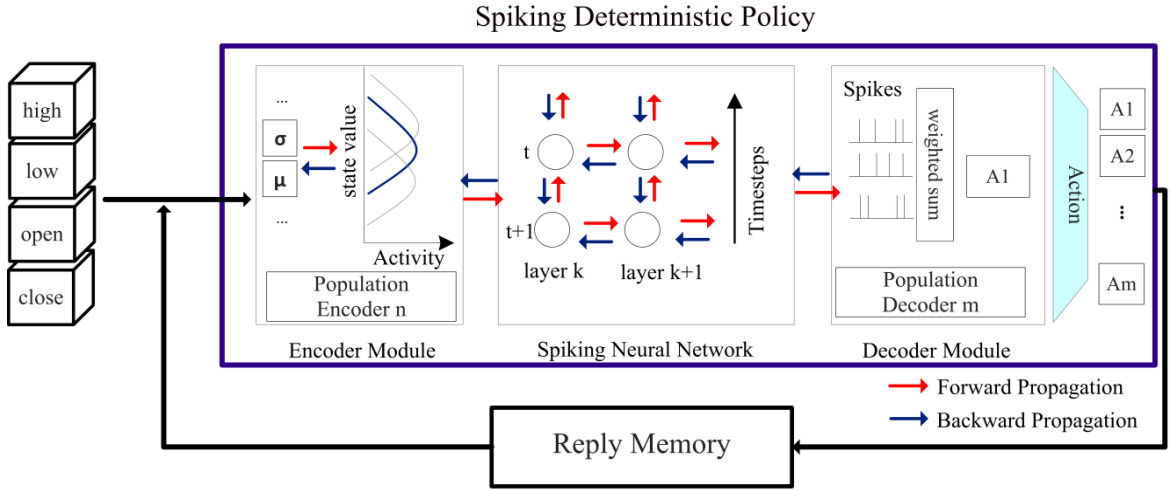

Fig. 1. training of multilayered SDP

Some SNN-based methods require much time for inference [25], which is especially problematic in PM for real-time decision-making. One possible way to solve this limitation is direct SNN training using gradient descent techniques such as spatiotemporal backpropagation (STBP) [26]. This method provides faster inference while displaying better performance for a wide range of issues [26], [27].

In this paper, we propose a deterministic spiking policy (SDP), a neuromorphic energy-efficient method based on population coding for optimal policy learning. We use the cryptocurrency market to evaluate the proposed method on 11 currencies with maximum volume. The unique feature of the proposed method is the ability to encode ample state space by population neurons with the ability to learn different parts, which effectively increases the network capacity. In this architecture, an SNN network is trained using gradient descent. For SDP training, we introduce a particular mode of STBP that allows SDP training on *Intel's Loihi*.

We evaluate the proposed method by comparing inference time and energy efficiency with the state-of-the-art methods and architectures. The proposed SDP on Loihi shows at least 186 times less energy consumption per inference than running other algorithms on CPU and at least 516x times less energy per inference than running other algorithms on GPU. while learning the optimal policy and performing better in terms of financial base parameters. In addition, the proposed algorithm provides 1.3x speed-up over the GPU and 2.0x speed-up over the CPU.

The remainder of this paper is organized as follows. In Section II, the PM problem is explained, and the proposed solution is described. Section III describes the evaluation methodology and the results. The last Section concludes and continues the future work.

## II. PROPOSED METHOD

### A. Problem Definition

In portfolio management, our goal is to maximize the profit by keeping trading risk low. We are presenting a deterministic spiking policy (SDP) learning algorithm (Fig. 1) that can learn an optimal policy by taking stock attributes that include *state = {$w_{t-1}$, close price, high price, low price, open price}* and a probability function that represents generates the amount of capital distribution among the assets at moment *t*. The mode includes the entry price, the lowest price, the highest price, the close price, and the weight of the capital division. On the other hand, the action consists of a weighting to place the capital in different stocks. This algorithm includes an SNN model that generates an action at time *t + 1* during a training process by receiving state at moment *t*.

In this case, the agent's goal is to maximize the final portfolio value ($\rho_f$) at the end of time $t_f + 1$. The agent has no control over the exact choice of the value $\rho_o$, and $t_f$ is the length of the portfolio management process interval. Hence, the goal is to maximize the average reward logarithm obtained according to (1) [12].

$$R = \frac{1}{t_f} \sum_{t=1}^{t_f+1} r_t$$

$$= \frac{1}{t_f} ln \frac{\rho_f}{\rho_o} = \frac{1}{t_f} \sum_{t=1}^{t_f+1} ln(\mu_t y_t . w_{t-1})$$

(1)

Where $\mu_t$ is transaction residual factor for real world market simulation, $y_t$ is price relative vector for $t^{th}$ and $w_{t-1}$ trading range are obtained by performing action on t-1. The reward function converges the SDP algorithm to achieve the goal during exploration, which facilitates training. For inference, we place the trained SDP algorithm on Loihi to obtain an action for the best portfolio value in the capital market. The training and inference process is described in the following subsections.

## B. Spiking deterministic policy (SDP):

The SDP network has an encoder module that uses it to spike real-world data. In this encoder module, we encode the dimensions of the action mode space with the input and output of spiking population neurons. For $i^{th}$ dimension of M-dimensional state $S_i$, $i \in \{1, ..., M\}$, we create a population of neurons, $E_i$, to encode the information. E is a Gaussian function with $\mu$ and deviation from the $\varepsilon$ criterion. The value of $\mu$ equals the equal distribution of state space, and $\varepsilon$ is a considerable predetermined value of non-zero population activity in all state spaces. The encoder module calculates the activity of the population E in two steps. In the first step, the state values are converted to stimulation strength for each population neuron, $A_E$, according to (2).

$$A_E = Exp\left(-\frac{1}{2}\cdot\left(\frac{s-\mu}{\sigma}\right)^2\right) \quad (2)$$

In the second step, the calculated $A_E$ is used to generate spike E neurons. There are two ways to do this: a) Probability encoding in which the spikes produced by all neurons at each timestep are probabilistically defined by $A_E$ and b) Deterministic encoding where the neurons inside E are simulated as one-step soft-reset Leaky-Integrateand-Fire (LIF) neurons in which $A_E$ plays the role of presynaptic inputs of neurons. Eq. (3) and (4) govern the dynamic of neurons.

$$V(t) = V(t-1) + A_E \quad (3)$$

$$o_k(t) = 1 \,\&\, V_k(t) = V_k(t) - (1-\varepsilon) \\ if \, V_k(t) > 1 - \varepsilon \quad (4)$$

Where k represents the index of each neuron in E and $\varepsilon$ is a small constant value. The output of the encoder module is input to the LIF neurons. LIF neurons have been used to build the SDP network (Fig. 1). These neurons update the $i^{th}$ state in time $t$ in two steps. In the first step, spike inputs are converted to synaptic current according to (5).

$$c_i(t) = d_c \cdot c_i(t-1) + \sum_j w_{ij} o_j(t) \quad (5)$$

Where $c$ is the synaptic current, $d_c$ is decay factor for the current, $w_{ij}$ is the binding weight of the $j^{th}$ of the presynaptic neuron, and $o_j$ is a value between 0 and 1, indicating the spike event j of the presynaptic neuron. According to the membrane voltage in (6) and (7), neurons are obtained through synaptic current in the second step. In this way, the neuron only produces a spike if the membrane voltage exceeds the threshold.

$$V_i(t) = d_v \cdot V_i(t-1) + c_i(t) \, if \, V_i(t-1) < V_{th} \quad (6)$$

$$o_i(t) = 1 \,\&\, V_i(t) = 0 \, otherwise \quad (7)$$

Where $V$ is the membrane voltage, $d_v$ is, decay factor for voltage, and $V_{th}$ is the spike fire threshold.

The decoder module converts the generated spike into three possible values for action. In the first step, after the T timestep, the spikes produced in the last layer are added together to obtain the firing rate after T. In the second step, $tempAction$ is obtained by weighting the sum $firingRate$ according to (8) and (9).

$$firingRate^{(i)} = \frac{sum \, spikes^{(i)}}{T} \quad (8)$$

$$tempAction^{(i)} = w_d^{(i)} \cdot firingRate^{(i)} + b_d^{(i)} \quad (9)$$

In the third step, action is brought using (10) (Algorithm 1).

$$a^{(i)} = \frac{tempAction^{(i)}}{\sum_{j=1}^{N} tempAction^{(j)}} \quad (10)$$

Where $N$ is the number of actions.

**Algorithm 1:** Forward propagation through SDP

**Output:** action $a^{(i)} \in [0,1] \, i \in \{1, ..., N\}, N = M + 1$, M = number of assets
**Require:** Maximum timestep $T$; Network depth $L$;
**Require:** $w^{(i)}$, $i \in \{1, ..., L\}$ the weight matrices
**Require:** $w_d^{(i)}$, $i \in \{1, ..., N\}$ the weight action matrices
**Require:** $b^{(i)}$, $i \in \{1, ..., L\}$ the bias parameters
**Require:** $w_d^{(i)}$, $i \in \{1, ..., N\}$ the action bias parameters
**Require:** $X^{(i)}$, $i \in \{1, ..., T\}$ the Input spike trains
**Require:** states $s$
Spike from Encoder Module $= X = Encoder(s, \mu, \sigma)$;
**for** $t = 1, ..., T$ **do**
$\quad o^{(t)(0)} = X^{(t)}$;
$\quad$ **for** $k = 1, ..., L$ **do**
$\quad\quad c^{(t)(k)} = d_c \cdot c^{(t-1)(k)} + w^{(k)} o^{(t)(k-1)} + b^{(k)}$;
$\quad\quad v^{(t)(k)} = d_v \cdot v^{(t-1)(k)} \cdot (1 - o^{(t-1)(k)}) + c^{(t)(k)}$;
$\quad\quad o^{(t)(k)} = Threshold(v^{(t)(k)})$;
$\quad$ **end**
**end**

$$SumSpike = \sum_{t=1}^{T} o^{(t)(L)}$$

**for** $t = 1, ..., N$ **do**
$\quad firingRate^{(i)} = \frac{sum \, spikes^{(i)}}{T};$
$\quad tempAction(i) = e^{w_d^{(i)} \cdot firingRate^{(i)} + b_d^{(i)}};$
**end**
**for** $t = 1, ..., N$ **do**
$\quad a^{(i)} = \frac{tempAction^{(i)}}{\sum_{j=1}^{N} tempAction^{(j)}}$
**end**

## C. training of SDP with Back Propagation:

We train the SDP network using the STBP algorithm to achieve optimal policy. The main STBP is limited to training networks with simple LIF neurons that have only one mode (voltage), but we use presented the STBP to train networks with LIF neurons in both current and voltage methods defined by (5) and (6) [18].

In this way, model training on Loihi with dual-state neurons is possible, because the threshold function that defines spike is immutable. The STBP algorithm requires a pseudo-gradient function to estimate the gradient of a spike. We use the rectangular function (defined in (11)) as the pseudo-gradient function, which gives us the best performance experimentally [26].

$$z(v) = \begin{cases} a_1 & if\ |v - V_{th}| < a_2 \\ 0 & otherwise \end{cases} \quad (11)$$

Where $z$ is pseudo-gradient, $a_1$ the amplifier of the gradient, and $a_2$ is threshold windows to pass the gradient.

At the end, SDP network action is obtained with forwarding propagation [18]. The parameters for the output i, $i \in \{I,..,M\}$ is updated independently according to (12).

$$\nabla_{w_d^{(i)}} = \nabla_{a^i} L . W_d^{(i)} . firingRate^{(i)}$$
$$\nabla_{b_d^{(i)}} L = \nabla_{a^{(i)}} L . W_d^{(i)} \quad (12)$$

The loss gradient concerning the SDP parameters for each k layer is obtained by adding the Backpropagation gradients from all timesteps according to (13).

$$\nabla_{w^{(k)}} L = \sum_{t=1}^{T} o^{(t)(k-1)} . \nabla_{c^{(t)(k)}} L, \nabla_{b^{(k)}} L = \sum_{t=1}^{T} \nabla_{c^{(t)(k)}} L \quad (13)$$

Table 1 PRICE DATA RANGE FOR DIFFERENT EXPERIMENTS

| **Experiment** | **Time range** | | |
|---|---|---|---|
| | *Training set* | *Back test set* | *Total* |
| 1 | 2016/08/01-2019/04/14 | 2019/04/14-2019/08/01 | 2016/08/01-2019/08/01 |
| 2 | 2017/08/01-2020/04/14 | 2020/04/14-2020/08/01 | 2017/08/01-2020/08/01 |
| 3 | 2018/08/01-2021/04/14 | 2021/04/14-2021/08/01 | 2018/08/01-2021/08/01 |

*D. SDP development on Loihi:*

The Loihi chip uses an 8-bit integer to store weights. To be able to map model weights to Loihi, we need to rescale them. Weights and threshold voltages for each layer are rescaled according to (14). The advantage of network training on Loihi is that all hyperparameters are the same values set at train time (Fig. 2).

$$r^{(k)} = \frac{w_{max}^{(k)(loihi)}}{w_{max}^{(k)}}$$
$$w^{(k)(loihi)} = round(r^{(k)} . w^{(k)}) \quad (14)$$
$$V_{th}^{(k)(loihi)} = round(r^{(k)} . V_{th})$$

Where $r^{(k)}$ is the rescale ratio of the k layer, $w_{max}^{(k)(loihi)}$ is the highest weight supported by Loihi, and $w_{max}^{(k)}$ is the highest weight of the trained k layer. $w^{(k)(loihi)}$ is the weight rescale on Loihi and $V_{th}^{(k)(loihi)}$ is the threshold voltage rescale on Loihi.

Table 2 HYPERPARAMETERS FOR TRAINING SDP

| Parameter | values |
|---|---|
| Neuron parameters (Vth, dc, dv) | 0.5, 0.5, 0.80 |
| Pseudo-gradient function parameters (a1, a2) | 9.0, 0.4 |
| Neurons per hidden layer for SDP | 128, 128 |
| Batchsize | 128 |
| Learning rate for training SDP | 10e-5 |

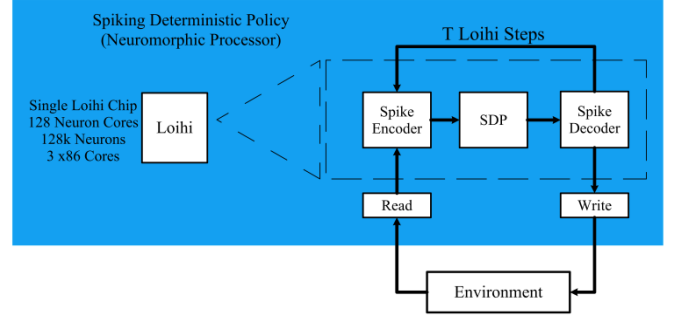

Fig. 2. SDP deployed on Intel's Loihi neuromorphic processor for energy-efficient and real-time portfolio management

III. EXPERIMENTS

We evaluate the proposed algorithm on cryptocurrency market data and design three experiments for evaluation algorithm. Each test consists of a portfolio of 11 cryptocurrencies with the highest trading volume in the last 30 days before the test data. This data is collected from polonix.com [28]. At the same time, we compare the performance, power consumption, and inference time of the investment strategies presented in this paper with popular investment strategies that have recently been compared. Since cryptocurrencies are traded 24 hours a day, all training and testing sets from 0:00 A-M Coordinated Universal Time (CUT). The duration of each experiment is three years. In each experiment, 80% of the collected data is considered for the training set and 20% for the algorithm test. The time frame of the training and test suite for each test is given in Table 1. To investigate the effect of forwarding propagation, we trained the SDP model for value T=5. The model hyperparameters used for training can be seen in Table 2.

*A. Performance Metrics*

This paper presents three performance metrics for evaluating portfolio management strategies. The first metric is the final accumulated Portfolio Value (fAPV). This measure shows the success rate of portfolio management in accumulation (APV), $p_t$, portfolio value over a period of time. fAPV is the APV at the end of the back-test experiment. fAPV is obtained by (15)

$$p_t = \frac{p_t}{p_0} \quad (15)$$

One of the disadvantages of the APV benchmark is that it does not take into account market risk. The second metric is the Sharp Ratio, which is used to consider risk. Sharp Ratio is obtained by (16)

$$Sharp\ Ratio = \frac{\mathbb{E}_t[p_t - p_f]}{\sqrt{var(p_t - p_f)}} \quad (16)$$

Where $p_t$ is the periodic return and $p_f$ is the risk-free asset rate.

The sharp ratio criterion causes portfolio values to be forgotten but does not consider upwards and downwards movement. For this purpose, we use the Maximum Draw-Down (MDD) metric. Mathematically, MDD is the most considerable loss from a peak to a through. MDD is obtained by (17)

$$MDD = max_{\tau > t} \frac{p_t - p_\tau}{p_t} \quad (17)$$

Table 3 PERFORMANCES OF THE PROPOSED METHOD AND BEST DRL METHOD BASED ON DNN AND SOME TRADITIONAL PORTFOLIO SELECTION STRATEGIES IN THREE DIFFERENT BACK-TEST EXPRIMENTS

| Strategy | Performance Matrix | | |
|---|---|---|---|
| | MDD | fAPV | Sharp Ratio |
| Experiment 1 | | | |
| SDP | **0.152** | **5.87e+7** | 0.245 |
| DRL[Jiang] | 0.159 | 4.41e+7 | **0.267** |
| ONS | 0.416 | 7.74e-01 | -0.008 |
| Best Stock | 0.627 | 1.580 | 0.014 |
| ANTICOR | 0.189 | 2.422 | 0.034 |
| M0 | 0.362 | 7.93e-01 | -0.005 |
| UCRP | 0.351 | 7.49e-1 | -0.014 |
| Experiment 2 | | | |
| SDP | 0.024 | **4.371** | 0.028 |
| DRL[Jiang] | **0.021** | 0.977 | -0.033 |
| ONS | 0.124 | 0.929 | -0.005 |
| Best Stock | 0.427 | 3.623 | **0.034** |
| ANTICOR | 0.784 | 0.222 | -0.086 |
| M0 | 0.189 | 1.240 | 0.017 |
| UCRP | 0.118 | 1.080 | 0.009 |
| Experiment 3 | | | |
| SDP | 0.253 | **2.009** | **0.037** |
| DRL[Jiang] | **0.249** | 1.760 | 0.031 |
| ONS | 0.365 | 0.925 | 0.001 |
| Best Stock | 0.511 | 8.380 | 0.036 |
| ANTICOR | 0.752 | 0.251 | -0.025 |
| M0 | 0.271 | 2.003 | 0.029 |
| UCRP | 0.231 | 1.840 | 0.033 |

*B. Discussion:*

Table 3 shows the fAPV, MDD, and Sharp Ratio performance scores for our proposed method and other portfolio management strategies. As shown in Table 3, the proposed method performed better than the other strategies in all experiments. This performance is to extract more features with a simple SNN model, which has resulted in faster and better training of the proposed model. Our proposed method is twice as good in fAPV, at least 20% in MDD, and at least 10% in Sharp Ratio. New methods such as Jiang do not perform well on new financial market data, despite a good performance from 2014 to 2017.

To compare and analyze the proposed solution with other recently proposed algorithms (Table 4), we consider the two parameters, average power consumption and inference speed, during the execution of the test data. The competitor algorithms run on Corei7-7500 CPU and Tesla K80 GPU processors, and the proposed algorithm runs on the Loihi chip. We use the tools that monitor onboard sensors to measure the power criterion of each device: powerstat for CPU, Nvidia-smi for GPU, and energy probe for Loihi.

Energy cost per inference is obtained by dividing the energy consumed per second by the number of inferences performed per second. Compared to our proposed algorithm, the algorithms running on CPU and GPU chips have more than 186 times more power consumption during high-speed inference. There is a trade-off for performance cost between SNN's with different timesteps, indicating that the larger the T, the better the performance cost, but the higher the energy cost.

Further reduction of energy costs and allowing portfolio management to be solved on devices such as mobile devices will reduce overall costs such as costs for powerful processing devices. While SNNs are more indicative of reduced energy consumption and sometimes cause a decline in performance [29]–[31], this study may be the first to show that an energy-efficient method is more accurate than other portfolio management algorithms during testing. In addition, the SNN's inherently noisy representation may have helped to escape the 'bad' local minimum [18].

In sum, all efforts have been made to implement a real-time and energy-efficient solution to the portfolio management problem. In future work, we can research the combination of DNN models with SNN in order to achieve high accuracy in solving the problem by reducing energy consumption.

Table 4 POWER PERFORMANCE ACROCC HARDWARE

| Algorithm | Device | Idle(w) | Dyn(w) | Inf/s | nJ/Inf |
|---|---|---|---|---|---|
| **DRL-Exp1** | CPU | 7.98 | 24.02 | 2.09 | 3835.85 |
| | GPU | 100.80 | 29.15 | 1.23 | 9165.32 |
| **SDP-Exp1** | Loihi (T=5) | 1.01 | 0.012 | 1.04 | **15.81** |
| **DRL-Exp2** | CPU | 9.09 | 22.91 | 1.60 | 2935.62 |
| | GPU | 100.25 | 29.66 | 1.09 | 8119.44 |
| **SDP-Exp2** | Loihi (T=5) | 1.01 | 0.011 | 0.82 | **15.72** |
| **DRL-Exp3** | CPU | 8.69 | 23.31 | 2.02 | 3706.38 |
| | GPU | 106.03 | 24.33 | 1.07 | 7998.76 |
| **SDP-Exp3** | Loihi (T=5) | 1.01 | 0.012 | 1.01 | **15.43** |

IV. CONCLUSION AND FUTURE WORK

This paper presents a neuromorphic framework combining lower power consumption and high robustness capabilities of SNN networks with DNN learning capabilities and tests them on the cryptocurrency market. While recent efforts have focused on architecture and restructuring that architecture we have introduced a new method using SNNs and changed models and their training. By changing the data structure, the training of our proposed model facilitated the training and transfer of important information during learning, which indicates a better feature extraction than other architectures in the past. As a result, an optimal and energy-efficient solution to the portfolio management problem was presented by implementing it on a neuromorphic Loihi processor.

Our method is by 186x more energy-efficient than previous algorithms on DNN compatible low-power devices. This performance advantage includes the asynchronous and event-based calculations provided by SNNs and the ability of our method to train SNN at low T values with negligible performance loss. This energy efficiency improvement enables us to solve the portfolio management problem on mobile and embedded devices with limited onboard resources.